\documentclass{article}

\usepackage[colorlinks]{hyperref}       % hyperlinks
\usepackage{amsfonts}       % blackboard math symbols
\usepackage{graphicx}
\usepackage{listings}
\usepackage{multirow}
\usepackage{multicol}
\usepackage{subcaption}
\usepackage{arxiv}
\usepackage{euscript}
\usepackage{amsmath} 
\usepackage{xspace}
\usepackage{caption}
\usepackage{subcaption}
\usepackage{booktabs}
\usepackage{textcomp}
\usepackage{amsmath} 
\usepackage{balance}
\usepackage{multirow}
\usepackage{multicol}
\usepackage{amssymb}
\usepackage{euscript}
\usepackage{array}
\usepackage[export]{adjustbox}
\newcolumntype{C}[1]{>{\centering\arraybackslash}p{#1}}

\begin{document}
\title{RaBiT: An Efficient  Transformer using Bidirectional Feature Pyramid Network with Reverse Attention for Colon Polyp Segmentation}
%
% \titlerunning{An Efficient  Transformer using Bidirectional Feature Pyramid Network}
\author{{Nguyen Hoang Thuan} \\
School of Information and Communication Technology\\
Hanoi University of Science and Technology\\
\texttt{hung.pv194069@sis.hust.edu.vn} \\
\And
{Nguyen Thi Oanh} \\
School of Information and Communication Technology\\
Hanoi University of Science and Technology\\
\texttt{oanhnt@soict.hust.edu.vn} \\
\And
{Nguyen Thi Thuy} \\
School of Science, Engineering and Technology\\
RMIT University\\
\texttt{thuy.nguyen43@rmit.edu.vn} \\
\And
{Stuart Perry} \\
School of Electrical and Data Engineering\\
University of Technology Sydney\\
\texttt{stuart.perry@uts.edu.au} \\
\And
{Dinh Viet Sang} \thanks{Corresponding author}\\
School of Information and Communication Technology\\
Hanoi University of Science and Technology\\
\texttt{sangdv@soict.hust.edu.vn} \\
%% \AND
%% Coauthor \\
%% Affiliation \\
%% Address \\
%% \texttt{email} \\
%% \And
%% Coauthor \\
%% Affiliation \\
%% Address \\
%% \texttt{email} \\
%% \And
%% Coauthor \\
%% Affiliation \\
%% Address \\
%% \texttt{email} \\
}

% Uncomment to override  the `A preprint' in the header
%\renewcommand{\headeright}{Technical Report}
%\renewcommand{\undertitle}{Technical Report}
\renewcommand{\shorttitle}{An Efficient  Transformer using Bidirectional Feature Pyramid Network}

\maketitle              % typeset the header of the contribution
\begin{abstract}
Automatic and accurate segmentation of colon polyps is essential for early diagnosis of colorectal cancer. Advanced deep learning models have shown promising results in polyp segmentation. However, they still have limitations in representing multi-scale features and generalization capability. To address these issues, this paper introduces RaBiT, an encoder-decoder model that incorporates a lightweight Transformer-based architecture in the encoder to model multiple-level global semantic relationships. The decoder consists of several bidirectional feature pyramid layers with reverse attention modules to better fuse feature maps at various levels and incrementally refine polyp boundaries. We also propose ideas to lighten the reverse attention module and make it more suitable for multi-class segmentation. Extensive experiments on several benchmark datasets show that our method outperforms existing methods across all datasets while maintaining low computational complexity. Moreover, our method demonstrates high generalization capability in cross-dataset experiments, even when the training and test sets have different characteristics.
\end{abstract}
\section{Introduction}
\label{sec:introduction}
% NEED PARAPHRASE
Colorectal cancer (CRC) is one of the leading causes of cancer-related death, with nearly one million deaths in 2020 \cite{siegel2023cancer}.
According to a longitudinal study \cite{corley2014adenoma}, for every 1\% increase in adenoma detection rate, the risk of colon cancer decreases by 3\%. 
Therefore, early detection and removal of polyps are essential for cancer prevention and treatment, and colonoscopy is considered the gold standard for detecting colon adenomas and colorectal cancer.
Attempts have been made to develop learning algorithms to deploy in computer-aided diagnostic (CAD) systems for the automatic detection and prediction of polyps, which could benefit clinicians in detecting lesions and lower the miss detection rate \cite{bisschops2019advanced,colonsegnet,ddanet}. 

Most segmentation models use an architecture similar to UNet \cite{unet}, where the encoder is typically constructed using Convolutional Neural Networks (CNNs). However, CNNs have several limitations despite achieving impressive performance in segmentation tasks. The main reason is that CNNs can only capture information at the local level while losing the context at the global level.
Furthermore, most existing encoder-decoder segmentation models often limit multi-scale feature fusion by a bottom-up aggregation path in the decoder. Multi-scale feature representation is one of the key factors contributing to the efficiency of a semantic segmentation model. HarDNet-MSEG \cite{hardnet_mseg}, BlazeNeo \cite{an2022blazeneo} used dense feature aggregation with dilated convolution layers to better capture global context information. Tan et al. \cite{tan2020efficientdet} showed that repeating the feature aggregation module enabled more efficient high-level feature representation. 

Attention mechanisms are a prevalent strategy in learning deep neural networks. To reduce computational costs while keeping the model focused on the relevant information, Oktay et al. \cite{oktay2018attention} presented an Attention Gate module for UNet. Later research, such as UNet++ \cite{unet++}, DDANet \cite{ddanet}, DoubleUnet \cite{doubleunet}, enhanced UNet by incorporating more skip connections to improve information flow during forward and backward computation. NeoUNet \cite{ngoc2021neounet} employed a lightweight CNN-based encoder with attention gates to achieve a high inference speed while maintaining a good segmentation accuracy. PraNet \cite{pranet} and CaraNet \cite{caranet} employed a Reverse Attention (RA) module \cite{ra}, which emphasizes the border between a polyp and its surroundings. The addition of attention modules is beneficial for most encoder-decoder neural networks.

Very recently, there has been an increasing interest in using Transformers \cite{khan2022transformers} for semantic image segmentation. Transformers utilize self-attention layers to represent the global relationship between input elements to better capture the global context of images. Successful models for polyp segmentation based on Transformers include TransUNet \cite{transunet} and TransFuse \cite{transfuse}. TransUNet combines a ViT encoder and a CNN decoder, which leads to high computational expenses. TransFuse introduces a parallel design that consists of a Transformer branch and a CNN branch. A so-called BiFusion module is then used to fuse the two branches at multiple levels. This complex architecture makes the network large and resource-hungry. 

\begin{figure*}[!ht]
\centering
\includegraphics[width=1.0\textwidth]{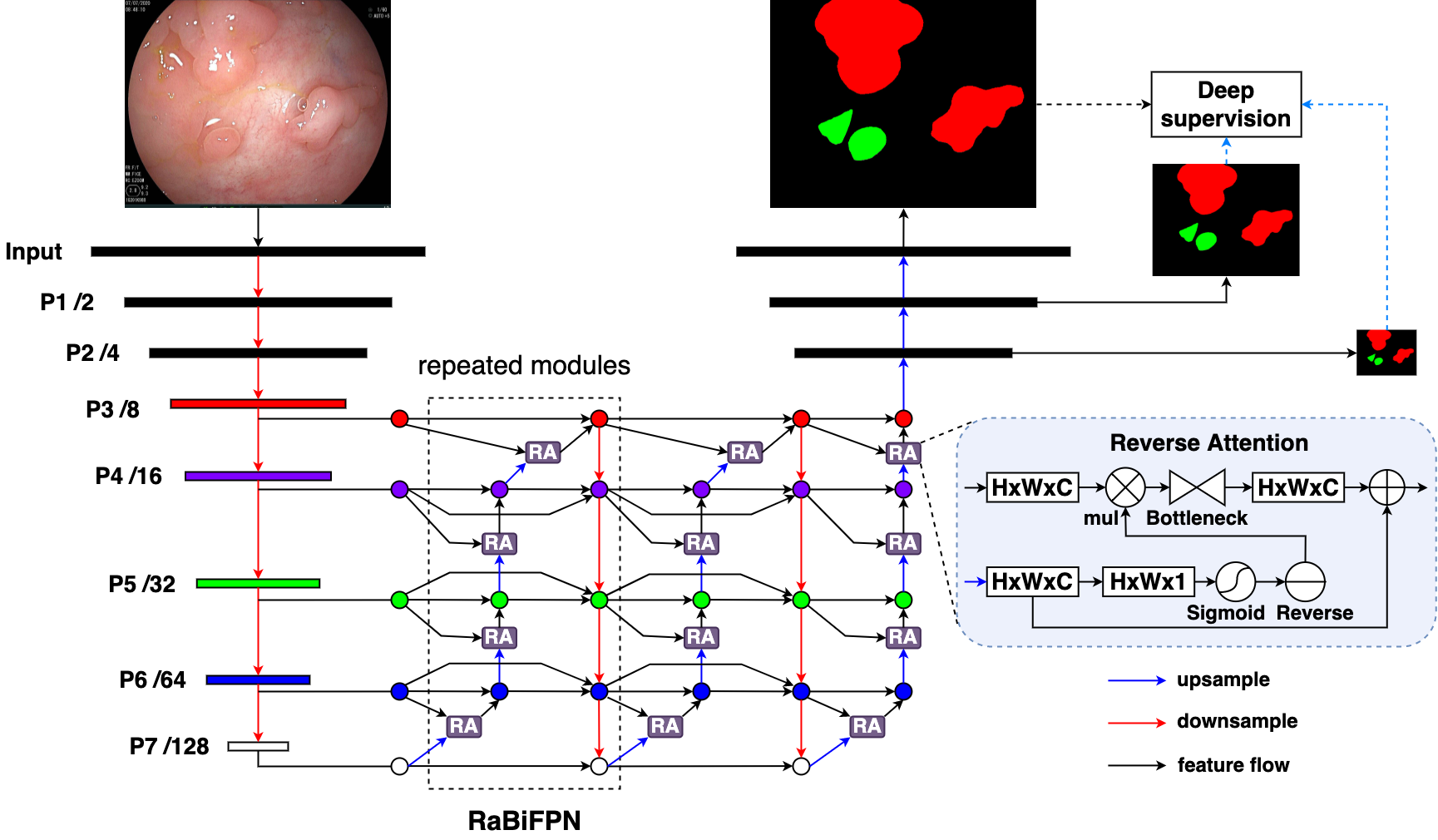} 
\caption{The overall architecture of our RaBiT contains two components: a lightweight MiT based encoder and a RaBiFPN decoder.} %See Section~\ref{sec:propose} for details.} 
\label{fig:overview}
\end{figure*}

Inspired by these approaches for modeling multi-scale and multi-level features, we propose a new Transformer-based network called RaBiT. Our approach differs from the available models in several ways. In brief, our main contributions are
fourfold. 1) In RaBiT, we design the encoder as a hierarchically structured lightweight Transformer for learning multi-scale features, and the decoder as a stack of bidirectional feature pyramid modules for learning from heterogeneous data containing feature maps extracted from encoder blocks at different scales and subregions; 2) A lightweight binary RA module using bottleneck layers is proposed to allow the repetition of this module several times in the stack of bidirectional feature pyramid layers to refine the segmentation boundaries of polyps incrementally; 3) A multi-class RA module is introduced, that is naturally suitable for multi-class segmentation; 4) An extensive set of experiments has been conducted on several standard benchmark datasets for polyp segmentation, and the results are compared with state-of-the-art methods.

The rest of the paper is organized as follows. %First, we briefly review related works in Section \ref{sec:related}. 
First, the RaBiT architecture is described in Section \ref{sec:propose}. Next, Section \ref{sec:experiment} presents our experiments and results. Finally, we conclude the paper and highlight future works in Section \ref{sec:conclude}.

\section{Method}
\label{sec:propose}
Our proposed network consists of a Transformer-based encoder and a decoder with a bidirectional feature pyramid network using consecutive RA modules. Therefore, we call our network RaBiT. The overall architecture of RaBiT is depicted in Fig.~\ref{fig:overview}. Details will be given in the following sections.

\subsection{Encoder}
Mix Transformer (MiT) \cite{segformer} provides a hierarchical feature representation solution, which was designed to generate CNN-like multi-scale features. MiT has a series of variants, from MiT-B1 to MiT-B5, with the same architecture but different sizes. For efficiency, MiT-B3 is used as the backbone of our network for all experiments in this paper. Let $P_i$ be the feature map with a resolution of $1/2^i$ of the input images. The MiT encoder has five feature levels $P_1 - P_5$. Inspired by \cite{tan2020efficientdet}, to capture more multi-scale features, we derive two new feature levels $P_6$ and $P_7$ from $P_5$. In order to obtain $P_6$, we pass the feature map $P_5$ through a 1x1 conv layer with 224 neurons and stride 1, followed by a batch norm and a 3x3 max pooling layer with stride 2. Next, a 3x3 max pooling layer is applied to the feature map $P_6$ to get the feature map $P_7$. Finally, all feature maps $P_3 - P_7$ are fed to the decoder of the networks. 

\subsection{Decoder}

The decoder of RaBiT is designed to leverage the strengths of both BiFPN \cite{tan2020efficientdet} and RA module \cite{pranet}. The original RA \cite{pranet} is employed in bottom-up direction to refine polyp boundaries, starting from the coarsest feature maps and progressing toward finer low-level ones. Inspired by BiFPN \cite{tan2020efficientdet}, we propose to iteratively refine the polyp boundaries using RA. After each round of bottom-up refinement, we continue to fuse information between feature levels in a top-down direction. We then repeat the bottom-up polyp boundary refinement process and continue this procedure several times. We refer to the combination of top-down feature aggregation and bottom-up boundary refinement as a \textbf{R}everse \textbf{A}ttention based \textbf{Bi}directional \textbf{F}eature \textbf{P}yramid \textbf{N}etwork (\textbf{RaBiFPN}) module, which can be iterated multiple times to form a stacked block. In this paper, we repeat the RaBiFPN module four times based on our ablation studies. All feature maps in the decoder are aggregated using fast normalized fusion \cite{tan2020efficientdet}, where fusion weights are learned by back-propagation during the training phase. Eventually, we incorporate a last refinement branch to produce the final prediction (see~Fig.~\ref{fig:overview}).

The original RA module \cite{pranet} takes an input and produces an output of size $H \times W \times 1$, and all the polyp boundary refinement steps are performed on single-channel feature maps. Compressing all feature maps into a single channel may result in the loss of critical information, reducing the effectiveness of multi-level feature fusion. In addition, single-channel processing is unsuitable for multi-class segmentation tasks, as each class must be processed separately in different channels.
To address these limitations, we introduce a modified RA module that operates on feature maps of size $H\times W\times C$, where $C=224$ is the number of channels in the last feature map ($P_7$). Before being fed into the decoder, we use a 3x3 conv layer with $C$ neurons to compress the higher resolution feature maps from $P_3$ to $P_5$ to the same depth of $C=224$. In contrast, the feature maps $P_6$ and $P_7$ have $C$ channels and can be fed directly to the decoder.

\begin{figure*}[!ht]
\centering
\includegraphics[width=0.8\textwidth]{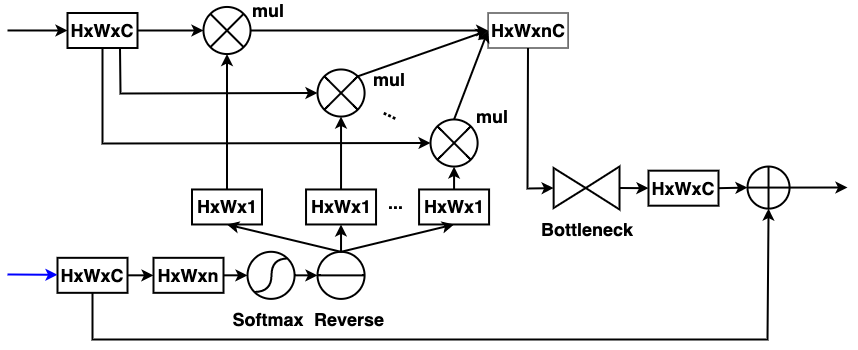} 
\caption{The architecture of Softmax RA module for multi-class segmentation.}
\label{fig:softmaxRA}
\end{figure*}

For binary segmentation, we use a 3x3 conv layer with one neuron to compress the input feature map to a single channel. We then apply a sigmoid function followed by a reverse operation to perform RA for a single polyp class (see Fig.~\ref{fig:overview}).
For multi-class segmentation, we use a 3x3 conv layer with $n$ neurons to compress the input feature map to $n$ channels, where $n$ is the number of classes. We then apply a softmax function to generate an attention map for each channel. Next, each attention map is reversed to produce a reverse attention map and element-wise multiplied with the input feature map. Finally, the resulting feature maps are concatenated to form a $H \times W \times nC$ volume (see~Fig.~\ref{fig:softmaxRA}).
To avoid parameter explosion when repeating the RA module multiple times in the network, we propose using a bottleneck module when performing the convolution operation to obtain the output feature map of the RA module. The bottleneck module consists of a 1x1 conv layer with $C/2$ neurons, followed by a 3x3 conv layer with $C/2$ neurons, and ends with a 1x1 conv layer with $C$ neurons.

\subsection{Loss Function}
For binary segmentation, %RaBiT uses a compound loss combining two types of loss function: the weighted focal loss and weighted IoU loss. 
we use a compound loss combining a weighted focal loss and a weighted IoU loss.
Similar to PraNet \cite{pranet}, the weighted losses are used to increase the weights of hard pixels. However, unlike PraNet, we utilize the focal loss \cite{focalloss} instead of BCE to deal with the imbalanced data problem in polyp segmentation. For multi-class polyp segmentation, the categorical cross-entropy loss is used.
Finally, we also apply deep supervision to multi-scale outputs to train the network, as shown in Fig.~\ref{fig:overview}. The final loss is the sum of all losses computed at the different output levels. Note that each output is upsampled back to the original size of the image's ground truth before the losses are calculated.

\section{Experiments}
\label{sec:experiment}

\subsection{Datasets}
We conducted experiments on five popular datasets for binary polyp segmentation: Kvasir \cite{kvasir}, CVC-Clinic DB \cite{cvc_clinic}, CVC-Colon DB \cite{cvc_colon}, CVC-T \cite{endo}, and ETIS-Larib Polyp DB \cite{etis_larib}. We then compared our approach with existing methods on two datasets for multi-class polyp segmentation,  NeoPolyp-Large and NeoPolyp-Small \cite{ngoc2021neounet}, where the challenging goal is not only to segment polyps from the background but also to determine benign and neoplastic polyps. 

\begin{table*}[!ht]
\caption{Performance comparison for binary polyp segmentation.}
\centering
{\renewcommand{\arraystretch}{1.0}
\begin{tabular}{c|cc|cc|cc|cc|cc}
\hline
\multirow{2}{*}{\textbf{Method}} & \multicolumn{2}{c|}{\textbf{Kvasir}} & \multicolumn{2}{c|}{\textbf{CVC-ClinicDB}} & \multicolumn{2}{c|}{\textbf{CVC-ColonDB}} & \multicolumn{2}{c|}{\textbf{CVC-T}} & \multicolumn{2}{c}{\textbf{ETIS-Larib}}  \\
\cline{2-11}
& mDice & mIoU               & mDice & mIoU                 & mDice & mIoU                & mDice & mIoU                   & mDice & mIoU              \\
\hline
UNet \cite{unet}           & 0.818 & 0.746 & 0.823 & 0.750 & 0.512 & 0.444 & 0.710 & 0.627  & 0.398 & 0.335 \\
UNet++ \cite{unet++}         & 0.821 & 0.743 & 0.794 & 0.729 & 0.483 & 0.410 & 0.707 & 0.624  & 0.401 & 0.344 \\
SFA   \cite{sfa}                      & 0.723 & 0.611 & 0.700 & 0.607 & 0.469 & 0.347 & 0.297 & 0.217 & 0.467 & 0.329\\
PraNet \cite{pranet}         & 0.898 & 0.840 & 0.899 & 0.849 & 0.709 & 0.640 & 0.871 & 0.797  & 0.628 & 0.567 \\
HarDNet-MSEG \cite{hardnet_mseg}  & 0.912 & 0.857 & 0.932 & 0.882 & 0.731 & 0.660 & 0.887 & 0.821  & 0.677 & 0.613 \\
TransUNet \cite{transunet}   & 0.913 & 0.857 & 0.935 & 0.887 & \underline{\textit{0.781}} & 0.699 & 0.893 & 0.824 & 0.731 & 0.660   \\
CaraNet \cite{caranet}          & 0.918 & 0.865 & 0.936 & 0.887 & 0.773 & 0.689 & \underline{\textit{0.903}} & \underline{\textit{0.838}} & \underline{\textit{0.747}} & \underline{\textit{0.672}}\\
%TransFuse-S \cite{transfuse}    & 0.918 & 0.868 & 0.918 & 0.868 & 0.773 & 0.696 & 0.902 & 0.833  & 0.733 & 0.659 \\
%TransFuse-L \cite{transfuse}    & 0.918 & 0.868 & 0.934 & 0.886 & 0.744 & 0.676 & 0.904 & 0.838  & 0.737 & 0.661 \\
TransFuse-L* \cite{transfuse}   & \underline{\textit{0.920}} & \underline{\textit{0.870}} & \textbf{0.942} & \textbf{0.897} & \underline{\textit{0.781}} & \underline{\textit{0.706}} & 0.894 & 0.826 & 0.737 & 0.663\\
%\hline

\textit{\textbf{RaBiT (Ours)}} & \textbf{0.927} & \textbf{0.879} & \underline{\textit{0.936}} & \underline{\textit{0.890}} & 
\textbf{0.824} & \textbf{0.749} & \textbf{0.905} & \textbf{0.839} & \textbf{0.823} & \textbf{0.747} \\[2pt]
\hline
\end{tabular}
}
\label{tab:sota}
\end{table*}

\begin{table*}[ht!]
\caption{Performance comparison for multi-class polyp segmentation.}
\centering
{\renewcommand{\arraystretch}{1.0}
\begin{tabular}{c|C{1.2cm} C{1.2cm}|C{1.2cm} C{1.2cm}|C{1.2cm} C{1.2cm} | c}
\hline
\multirow{2}{*}{\textbf{Method}} & \multicolumn{6}{c|}{\textbf{NeoPolyp-Large}} & \textbf{NeoPolyp-Small} \\
\cline{2-8}
& Dice$_{seg}$ & IoU$_{seg}$              & Dice$_{non}$ & IoU$_{non}$                & Dice$_{neo}$ & IoU$_{neo}$ & Kaggle score  \\
\hline
UNet \cite{unet}        & 0.785                                   & 0.646                                  & 0.525                                   & 0.356                                  & 0.773                                   & 0.631                                     & -                  \\
ColonSegNet \cite{colonsegnet}       & 0.738                                   & 0.585                                  & 0.505                                   & 0.338                                  & 0.732                                   & 0.577                                     & -                  \\
DDANet \cite{ddanet}        & 0.813                                   & 0.684                                  & 0.578                                   & 0.406                                  & 0.802                                   & 0.670                                     & -                  \\
DoubleU-Net \cite{doubleunet}    & 0.837                                   & 0.720                                  & 0.621                                   & 0.450                                  & 0.832                                   & 0.712                                     & -                 \\

HarDNet-MSEG\cite{hardnet_mseg} &  0.883 & 0.791 & 0.659 & 0.492 & 0.869 & 0.769 & -   \\
% SegFormer-Uper-ARA  & 0.921 & 0.871 & 0.928 & 0.881 & \textbf{0.815} & \textbf{0.737} & \textbf{0.912} & \textbf{0.848}  & 0.787 & 0.708 \\
PraNet \cite{pranet} & 0.895 & 0.811 &0.705 & 0.544 & 0.873 & 0.775 & - \\
BlazeNeo-DHA \cite{an2022blazeneo} & 0.904 & 0.825 &0.717 & 0.559 & 0.885 & 0.792 & 0.788  \\

NeoUNet \cite{ngoc2021neounet} &  \underline{\textit{0.911}} & \underline{\textit{0.837}} & \underline{\textit{0.720}} & \underline{\textit{0.563}} & \underline{\textit{0.889}} & \underline{\textit{0.800}} & \underline{\textit{0.807}} \\

\textit{\textbf{RaBiT (Ours)}}  & \textbf{0.940} & \textbf{0.886} & \textbf{0.765} & \textbf{0.619} & \textbf{0.917} & \textbf{0.846} & \textbf{0.859}  \\[2pt]
\hline
\end{tabular}
}
\label{tab:neo-sota}
\end{table*}

\subsection{Experiment Settings and Metrics}
\label{subsec:settings}
We implemented RaBiT using PyTorch. %\cite{...}. %For a fair comparison, we used the same parameters as \cite{segformer} for the MiT backbone: kernel size $K=7$, stride $S=4$, padding size $P=3$, and $K=3, S=2, P=1$ to produce features with the same size as the non-overlapping process. 
%Based on experiments in \cite{focalloss,f3net}, we use $\lambda=5, \alpha=0.25$ and $\gamma=2$ for the losses in Eq.~(\ref{eq:wfocal}) and Eq.~(\ref{eq:wIoU}). 
We trained the networks on a machine with 64GB RAM and an RTX 3090 GPU. Input images were resized to $384\times384$. We employed a multi-scale training strategy with scales of \{256, 384, 512\}. 
Standard augmentation techniques were used, such as horizontal/vertical flip, random rotation of $90^{\circ}$, gaussian blur, color jitter, and random crop.
We used the Adam optimizer and cosine annealing scheduler with a learning rate of 1e-4. RaBiT was trained in 30 epochs with a batch size of 8, and the last checkpoint was used for evaluation. We trained RaBiT five times, and results (except for the last column in Table~\ref{tab:neo-sota}) were averaged over five runs for all experiments.

We set up two groups of experiments to evaluate our method. 1) Binary polyp segmentation:  We used the same split as suggested in \cite{pranet}, where $90\%$ of the Kvasir and ClinicDB datasets were used for training. The remaining images in the Kvasir and CVC-ClinicDB datasets were used for testing, and all from CVC-ColonDB, CVC-T, and ETIS-Larib were used for cross-dataset testing; 2) Multi-class polyp segmentation: we used the NeoPolyp-Large dataset and followed the data split in \cite{ngoc2021neounet}. We also used a reduced version called NeoPolyp-Small \cite{neopolyp-small} for comparing different methods.  
\begin{figure*}[!ht]
\centering
\includegraphics[width=1.0\textwidth]{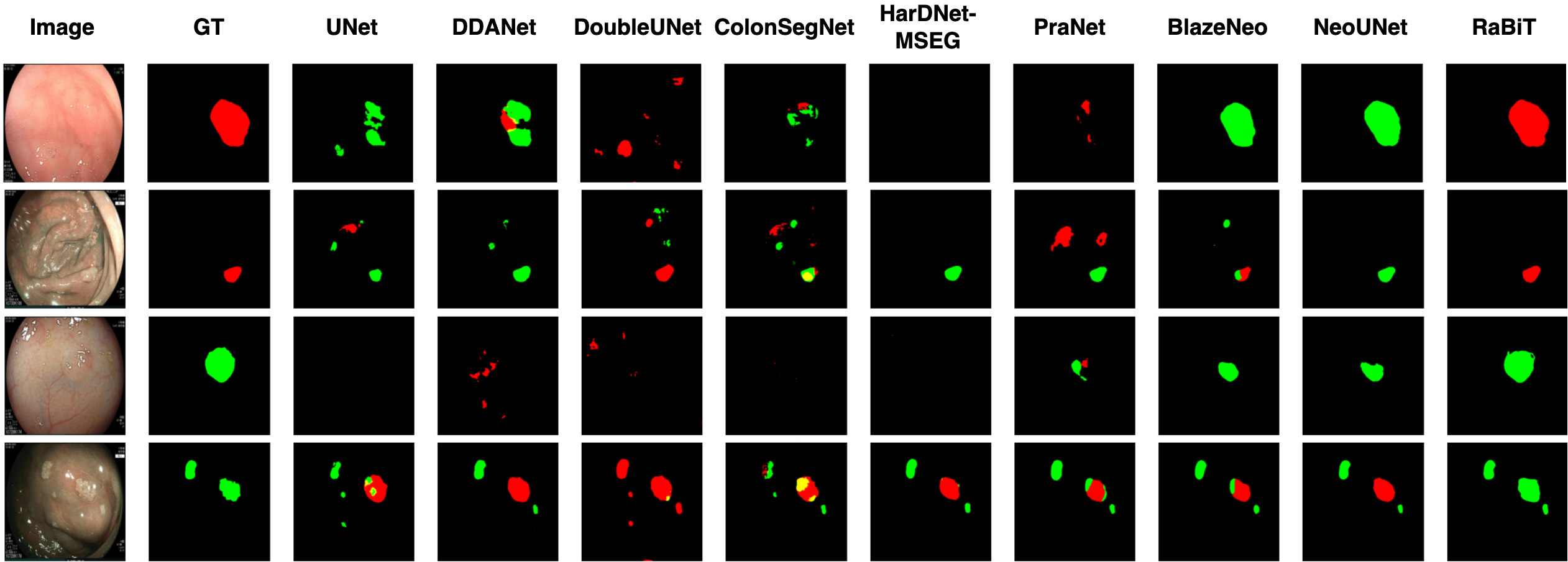}
\caption{Qualitative result comparison on the Neo-Large dataset.}
\label{fig:neo}
\end{figure*}

Similar to \cite{pranet}, we employ the mDice and mIoU metrics for quantitative evaluation of binary polyp segmentation. For multi-class polyp segmentation, we calculate micro mDice and micro mIoU for neoplastic, non-neoplastic, and generic polyps, respectively, as suggested in \cite{ngoc2021neounet}. Besides, since the test set of the NeoNeoPolyp-Small dataset is not public, we use the Kaggle mDice score described in \cite{neopolyp-small} to compare different methods on the Kaggle leaderboard.

\subsection{Comparison with the State-of-the-art (SOTA) Methods}
The performance comparison for binary polyp segmentation is described in Table~\ref{tab:sota}. RaBiT generally outperforms the benchmark models on most datasets.  RaBiT outperforms the second-best TransFuse-L* by $4.3\%$ in mDice and $4.3\%$ in mIoU on CVC-ColonDB. %containing the most images. 
Notably, RaBiT achieves a significant improvement of $7.6\%$ in mDice and $7.5\%$ in mIoU compared to the second-best CaraNet \cite{caranet} on ETIS-Larib. The efficacy of the RaBiFPN module within RaBiT appears to be well-suited for the high-resolution images in ETIS-Larib. However, RaBiT has roughly $0.7\%$ lower metric values than the best TransFuse-L* on CVC-ClinicDB, which contains low-resolution images.

Table~\ref{tab:neo-sota} shows the performance comparison of SOTA methods on three-class polyp segmentation datasets. RaBiT outperforms the second-best method, NeoUNet \cite{ngoc2021neounet}, for every metric by a large margin of about 3\%-6\%. % Specifically, in Dice$_{seg}$ and IoU$_{seg}$, which only evaluate the polyp segmentation and ignore the polyp classification task, RaBiT showed an improvement of  $2.9\%$ and $4.9\%$, respectively. RaBiT also outperforms the NeoUNet \cite{ngoc2021neounet} by $4.5\%$ and $5.6\%$ in Dice$_{non}$ and IoU$_{non}$ respectively, which are the metrics for evaluating the non-neoplastic polyp segmentation capability. With Dice$_{neo}$ and IoU$_{neo}$, RaBiT showed an improvement of $2.8\%$ and $4.6\%$ compared to NeoUNet \cite{ngoc2021neounet}. RaBiT also outperformed the second-best method, NeoUNet, by a large margin according to the Kaggle score on the NeoPolyp-Small dataset. This result shows that RaBiT proves not only effective on binary but also on multi-class segmentation tasks.
Fig.~\ref{fig:neo} shows the qualitative comparison between our approach and other SOTA methods on the NeoPolyp-Large dataset. One can see that RaBiT yields better segmentation results than other SOTA methods in many challenging cases.

\subsection{Learning Capability}
To better evaluate the learning capability of our method, we conducted 5-fold cross-validation on the CVC-ClinicDB and Kvasir datasets. Each dataset is divided into five equal folds. Each run uses one fold for testing and four remaining folds for training. Table~\ref{tab:kfold} describes the comparison results for this experiment. The average value and standard deviation are used as metrics to prove the models' stability. Our RaBiT not only outperforms all other state-of-the-art models in mDice, mIoU, precision, and recall on both datasets but also is the most stable model on both datasets, achieving the lowest standard deviation for each evaluation metric.

\begin{figure*}[ht!]
\centering
\includegraphics[width=1.0\textwidth]{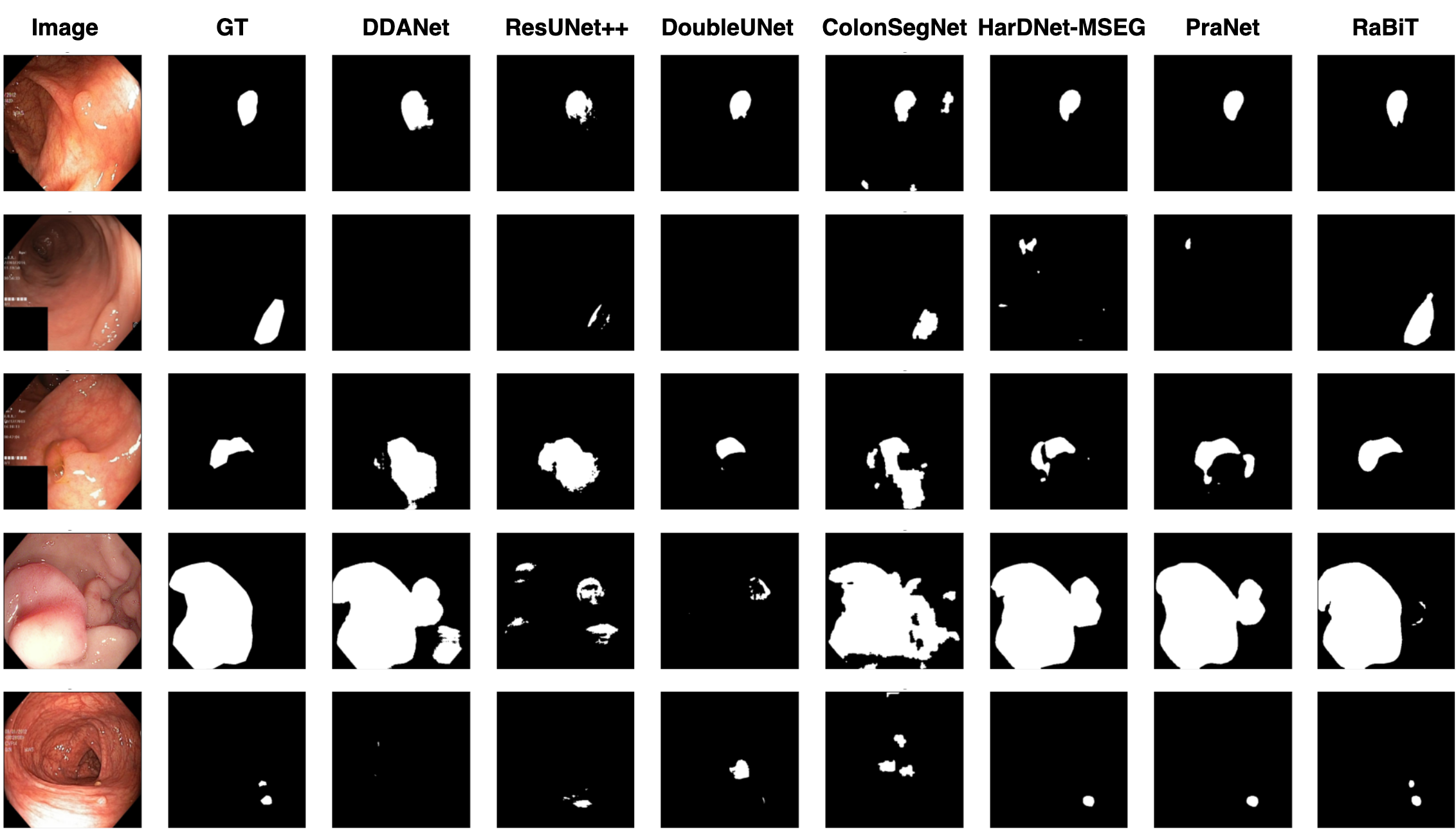}
\caption{Qualitative result comparison of different models trained on the first fold of the 5-fold cross-validation on the Kvasir dataset.}
\label{fig:kvasir_kfold}
\end{figure*}

Qualitative results for this experiment are shown in Fig.~\ref{fig:kvasir_kfold}. RaBiT demonstrates much fewer wrongly predicted pixels in segmentation results than other models.
\begin{table*}[ht!]
\caption{Performance comparison of different methods on 5-fold cross-validation of the CVC-ClinicDB and Kvasir datasets. All results are averaged over five folds.}
\centering
{\renewcommand{\arraystretch}{1.2}
\begin{tabular}{c|c|cccc}
\hline
\textbf{Dataset} & \textbf{Method} & \textbf{mDice} & \textbf{mIoU}  & \textbf{Recall} & \textbf{Precision}  \\
\hline
\hline
\multirow{9}{*}{\rotatebox[origin=c]{90}{ClinicDB}} 
& UNet \cite{unet}                   & -          & 0.792              & -          & -           \\
& MultiResUNet \cite{multiresunet}& -          & 0.849              & -          & -           \\
& ResUNet++ \cite{resunet++}  & $0.815 \pm 0.018$ & $0.736 \pm 0.017$ & $0.832\pm 0.018$ &  $0.830 \pm 0.020$ \\
& DoubleUNet \cite{doubleunet} & $0.920 \pm 0.018$ & $0.866 \pm 0.025$ & $0.922 \pm 0.027$ & $0.928 \pm 0.017$ \\
& DDANet \cite{ddanet} & $0.860 \pm 0.014$ & $0.786 \pm 0.017$ & $0.858 \pm 0.023$ & $0.892 \pm 0.014$ \\
& ColonSegNet \cite{colonsegnet} & $0.817 \pm 0.020$ & $0.873 \pm 0.024$ & $0.926 \pm 0.025$ & $0.933 \pm 0.014$ \\
& HarDNet-MSEG \cite{hardnet_mseg} & $0.923 \pm 0.020$ & $0.873 \pm 0.024$ & $0.926 \pm 0.025$ & $0.933 \pm 0.014$ \\
& PraNet \cite{pranet} & \underline{\textit{$0.933 \pm 0.012$}} & \underline{\textit{$0.884 \pm 0.015$}} & \underline{\textit{$0.940 \pm 0.005$}} & \underline{\textit{$0.937 \pm 0.016$}} \\
%& AG-CUResNeSt-101 \cite{agcuresnest} & $0.946 \pm 0.010$ & $0.902 \pm 0.015$ & $0.953 \pm 0.013$ & \textbf{0.944 $\pm$ 0.009} \\
%\cline{2-6}
& \textit{\textbf{RaBiT (Ours)}} & \textbf{0.951 $\pm$ 0.002} & \textbf{0.911 $\pm$ 0.002} & \textbf{0.949 $\pm$ 0.003} & \textbf{0.956 $\pm$ 0.002} \\
[2pt]
\hline
\hline
\multirow{8}{*}{\rotatebox[origin=c]{90}{Kvasir}} 
& UNet \cite{unet}                 & $0.708\pm0.017$          & $0.602\pm0.010$         & $0.805\pm0.014$  & $0.716\pm0.020$      \\
& ResUNet++ \cite{resunet++}            & $0.780\pm0.010$          & $0.681\pm0.008$         & $0.834\pm0.010$  & $0.799\pm0.010$      \\
& DoubleUNet \cite{doubleunet} & $0.879 \pm 0.018$ & $0.816 \pm 0.026$ & $0.902 \pm 0.027$ & $0.894 \pm 0.039$ \\
& DDANet \cite{ddanet} & $0.860 \pm 0.005$ & $0.791 \pm 0.004$ & $0.876 \pm 0.015$ & $0.892 \pm 0.018$ \\
& ColonSegNet \cite{colonsegnet} & $0.676 \pm 0.037$ & $0.557 \pm 0.040$ & $0.731 \pm 0.088$ & $0.730 \pm 0.080$ \\
& HarDNet-MSEG \cite{hardnet_mseg} & \underline{\textit{$0.889 \pm 0.011$}} & \underline{\textit{$0.831 \pm 0.011$ }}& $0.892 \pm 0.015$ &\underline{\textit{ $0.926 \pm 0.014$ }}\\
& PraNet \cite{pranet} & $0.883\pm0.020$          & $0.822\pm0.020$         & \underline{\textit{$0.897\pm0.020$}}  & $0.906\pm0.010$      \\
%& AG-CUResNeSt-101 \cite{agcuresnest}) & $0.912\pm0.010$ & $0.860\pm0.011$       & $0.923\pm0.009$  & \textbf{0.927 $\pm$ 0.014}      \\
%\cline{2-6}

& \textit{\textbf{RaBiT (Ours)}} & \textbf{0.921 $\pm$ 0.005} & \textbf{0.873 $\pm$ 0.005} & \textbf{0.927 $\pm$ 0.01} & \textbf{0.938 $\pm$ 0.006} \\[2pt]
\hline
\end{tabular}
}
\label{tab:kfold}
\end{table*}

\subsection{Generalization Capability}
To better evaluate the generalization capability, we conducted additional experiments with the following configurations:
\begin{itemize}
\item Configuration 1: CVC-ColonDB and ETIS-Larib for training, CVC-ClinicDB for testing;
\item Configuration 2: CVC-ColonDB for training, CVC-ClinicDB for testing;
\item Configuration 3: CVC-ClinicDB for training, ETIS-Larib for testing.
\end{itemize}

\begin{figure*}[ht!]
\centering
\includegraphics[width=1.0\textwidth]{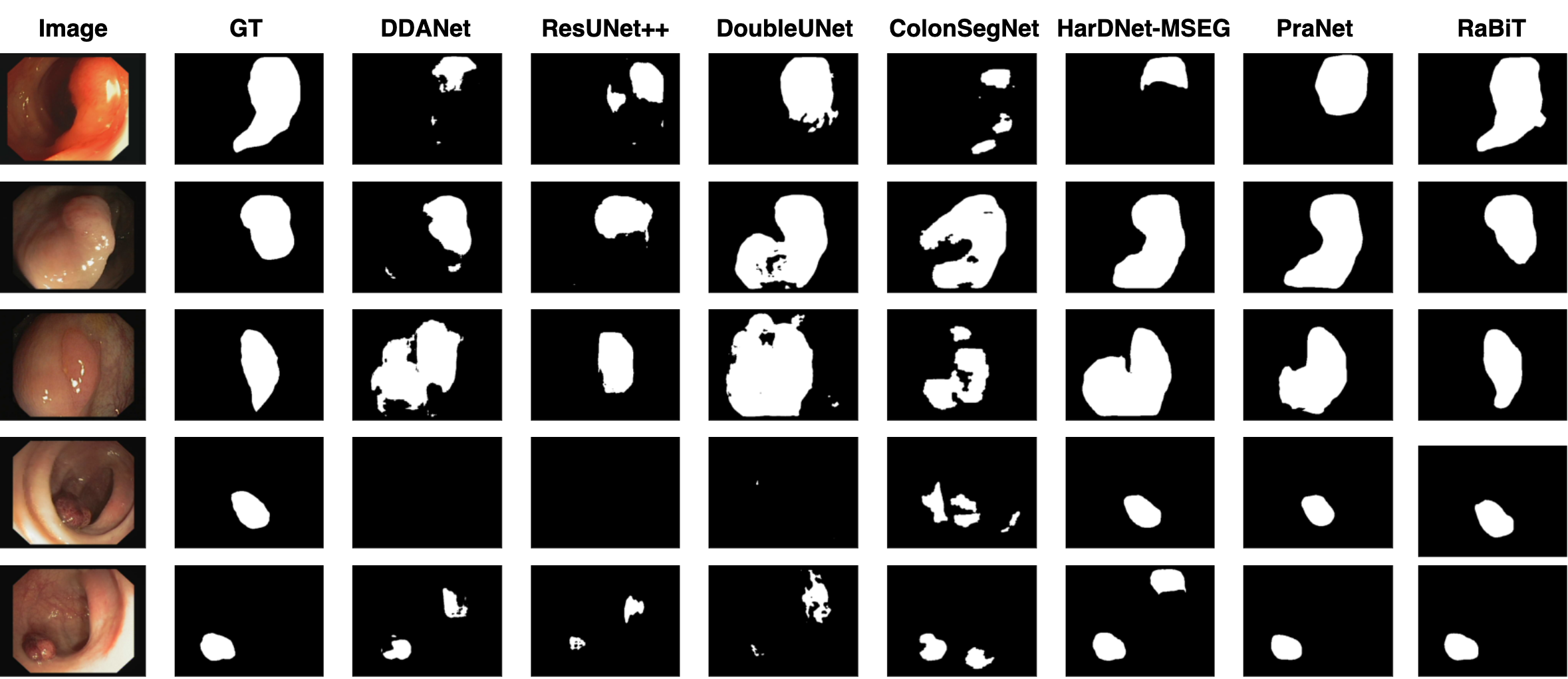}
\caption{Qualitative result comparison using CVC-Colon for training and CVC-Clinic for testing.}
\label{fig:colon_clinic}
\end{figure*}

Table~\ref{tab:cross-dataset} shows the comparison results for these experiments. Overall, RaBiT significantly outperforms benchmark models on every cross-dataset metric. For the first configuration, RaBiT outperforms the second-best PraNet by $6.8\%$ on mDice and $8.3\%$ on mIoU. In the second configuration, RaBiT continues to achieve a $9.7\%$ improvement in recall, $5.8\%$ improvement in mDice, and $6.8\%$ improvement in mIoU over PraNet. Especially for the third configuration, RaBiT again shows its suitability to the ETIS-Larib dataset achieving a $15\%$ improvement in mDice, $14.5\%$ in mIoU, $10\%$ in recall and $19.5\%$ in precision over the second-best HarDNet-MSEG. These are highly significant improvements, showing that our RaBiT can generalize well to new unseen data. Some result samples for this experiment are shown in Fig.~\ref{fig:colon_clinic}. Similar to Fig.~\ref{fig:kvasir_kfold}, one can see that RaBiT produces better segmentation results than other state-of-the-art.

\begin{table*}[!ht]
\caption{Performance comparison of different methods on cross-dataset configurations. All results are averaged over five runs.}
\centering
{\renewcommand{\arraystretch}{1.2}
\begin{tabular}{cc|c|c|C{1.2cm} C{1.2cm} C{1.2cm} c}
\hline
\multicolumn{2}{c|}{\textbf{Training}} & \textbf{Test} & \textbf{Method} & \textbf{mDice} & \textbf{mIoU}  & \textbf{Recall} & \textbf{Precision}  \\
\hline
\hline
\multirow{7}{*}{\rotatebox[origin=c]{90}{CVC-ColonDB}} & \multirow{7}{*}{\rotatebox[origin=c]{90}{+ ETIS-Larib}}  &  \multirow{7}{*}{\rotatebox[origin=c]{90}{CVC-ClinicDB}} 
& ResUNet++ \cite{resunet++} & 0.406 & 0.302 & 0.481 & 0.496 \\
&& & ColonSegNet \cite{colonsegnet} & 0.427 & 0.321 & 0.529 & 0.552 \\
&& & DDANet \cite{ddanet} & 0.624 & 0.515 & 0.697 & 0.692 \\
&& & DoubleUNet \cite{doubleunet} & 0.738 &  0.651 & 0.758 & 0.824 \\
&& & HarDNet-MSEG \cite{hardnet_mseg} & 0.765 &  0.681 & 0.774 & 0.863 \\
&& & PraNet \cite{pranet} & \underline{\textit{0.779 }}&  \underline{\textit{0.689}} & \underline{\textit{0.832}} & \underline{\textit{0.812}} \\
%& & AG-CUResNeSt-101 \cite{agcuresnest} & 0.833 & 0.754 & 0.840 & 0.883  \\
%\cline{3-7}

&& & \textit{\textbf{RaBiT (Ours)}}     &   \textbf{0.847}  &   \textbf{0.772}  &   \textbf{0.894}   &  \textbf{0.857} \\[2pt]
\hline
\hline
\multirow{9}{*}{\rotatebox[origin=c]{90}{CVC-ColonDB}} & &  \multirow{9}{*}{\rotatebox[origin=c]{90}{CVC-ClinicDB}} 
& ResUNet++ \cite{resunet++} & 0.339 & 0.247 & 0.380 & 0.484 \\
&& & DoubleUNet \cite{doubleunet} & 0.441 &  0.375 & 0.423 & 0.639 \\
&& & DDANet \cite{ddanet} & 0.476 & 0.370 & 0.501 & 0.644 \\
%& ResNet50-Mask-RCNN \cite{maskrcnn} & 0.639    & 0.560        & 0.648  & 0.710      \\
&& & ResNet101-Mask-RCNN \cite{maskrcnn} & 0.641 & 0.565        & 0.646  & 0.725      \\
&& & ColonSegNet \cite{colonsegnet} & 0.582 &    0.268 & 0.511 & 0.460 \\
&& & HarDNet-MSEG \cite{hardnet_mseg} & 0.721 &  0.633 & 0.744 & 0.818 \\
&& & PraNet \cite{pranet} & \underline{\textit{0.738 }}& \underline{\textit{0.647}} & \underline{\textit{0.751}} & \textbf{0.832} \\
%& & AG-CUResNeSt-101 \cite{agcuresnest}   & 0.771  & 0.686        & 0.793  & 0.830  \\
%\cline{3-7}

&& & \textit{\textbf{RaBiT (Ours)}} & \textbf{0.796}   &   \textbf{0.715} & \textbf{0.848}  & \underline{\textit{0.820}}\\[2pt]
\hline
\hline
\multirow{10}{*}{\rotatebox[origin=c]{90}{CVC-ClinicDB}} & &  \multirow{10}{*}{\rotatebox[origin=c]{90}{ETIS-Larib}} 
& ResUNet++ \cite{resunet++} & 0.211 & 0.155 & 0.309 & 0.203 \\
&& & ColonSegNet \cite{colonsegnet} & 0.217 &    0.110 & 0.654 & 0.144 \\
%&&&  ResNet50-Mask-RCNN \cite{maskrcnn} & 0.501   & 0.412         & 0.546  & 0.573      \\
&& & DDANet \cite{ddanet} & 0.400 & 0.313 & 0.507 & 0.464 \\
&& & ResNet101-Mask-RCNN \cite{maskrcnn} & 0.565   & 0.469         & 0.565  & 0.639      \\
&& & DoubleUNet \cite{doubleunet} & 0.588   & 0.500         & 0.689  & 0.599      \\
&& & PraNet \cite{pranet} & 0.631 &    0.555 &  \underline{\textit{0.762}}  &    0.597\\
&& & HarDNet-MSEG \cite{hardnet_mseg} & \underline{\textit{0.659}} &  \underline{\textit{0.583}} & 0.676 & \underline{\textit{0.705}} \\
%& & AG-CUResNeSt-101 \cite{agcuresnest} & 0.701  & 0.613         & 0.755  & 0.693      \\
%\cline{3-7}

&& & \textit{\textbf{RaBiT (Ours)}}  & \textbf{0.809}    &   \textbf{0.728} & \textbf{0.776}  & \textbf{0.900}\\[2pt]
\hline
\end{tabular}
}
\label{tab:cross-dataset}
\end{table*}

\subsection{Adaptation Capability to other Medical Image Segmentation Tasks}
We demonstrate the generalization capability of our method to other medical image analysis tasks by conducting an experiment on the ISIC 2018 dataset for skin lesion segmentation. Results are shown in Table~\ref{tab:isic}. One can see that our RaBiT yields superior performance compared to other state-of-the-art methods. This suggests that our RaBiT can serve as a strong baseline for other medical image segmentation tasks.
\begin{table*}[!ht]
\caption{Performance comparison of different methods on ISIC-2018 dataset.}
\centering
{\renewcommand{\arraystretch}{1.2}
\begin{tabular}{c | C{1.2cm} c C{1.2cm} c }
\hline

%\cline{2-7}
\textbf{Method}& \textbf{mDice} & \textbf{Precision} & \textbf{Recall} & \textbf{Accuracy}         \\
\hline
\hline
UNet \cite{unet} &  0.647 & 0.779 & 0.708 & 0.890   \\
%SegFormer-Uper & 0.921 & 0.869 & 0.928 & 0.881 & 0.795 & 0.718 & 0.904 & 0.838  & 0.782 & 0.704 \\
Attention UNet\cite{oktay2018attention} &  0.665 & 0.787 & 0.717 & 0.897   \\
% SegFormer-Uper-ARA  & 0.921 & 0.871 & 0.928 & 0.881 & \textbf{0.815} & \textbf{0.737} & \textbf{0.912} & \textbf{0.848}  & 0.787 & 0.708 \\
R2U-Net \cite{alom2018recurrent} & 0.679 & 0.741 &0.792 & 0.880  \\
Attention R2U-Net & 0.691 & 0.822 &0.726 & 0.904   \\
BCDU-Net(d=3) \cite{azad2019bi} & 0.851 & 0.928 &0.785 & 0.937   \\
MCGU-Net(d=3)\cite{asadi2020multi} &  \underline{\textit{0.895}} & \textbf{0.947} & \underline{\textit{0.848}} & \underline{\textit{0.955}}  \\

\textit{\textbf{RaBiT (Ours)}}  & \textbf{0.904} & \underline{\textit{0.917}} & \textbf{0.916} & \textbf{0.964}   \\[2pt]
\hline
\end{tabular}
}
\label{tab:isic}
\end{table*}

\subsection{Complexity Comparison}
Table~\ref{tab:complexity} compares RaBiT with other benchmark models in terms of size and computational complexity. Our RaBiT obtains competitive size and computational complexity compared to the most lightweight CNN-based models such as PraNet \cite{pranet}, and HarDNet-MSEG \cite{hardnet_mseg}. RaBiT is larger than most CNN-based neural networks but still more efficient than other Transformer-based methods in terms of GFlops.

\begin{figure}[!ht]
     \centering
     \begin{subfigure}[b]{0.495\textwidth}
         \centering
         \includegraphics[width=\textwidth]{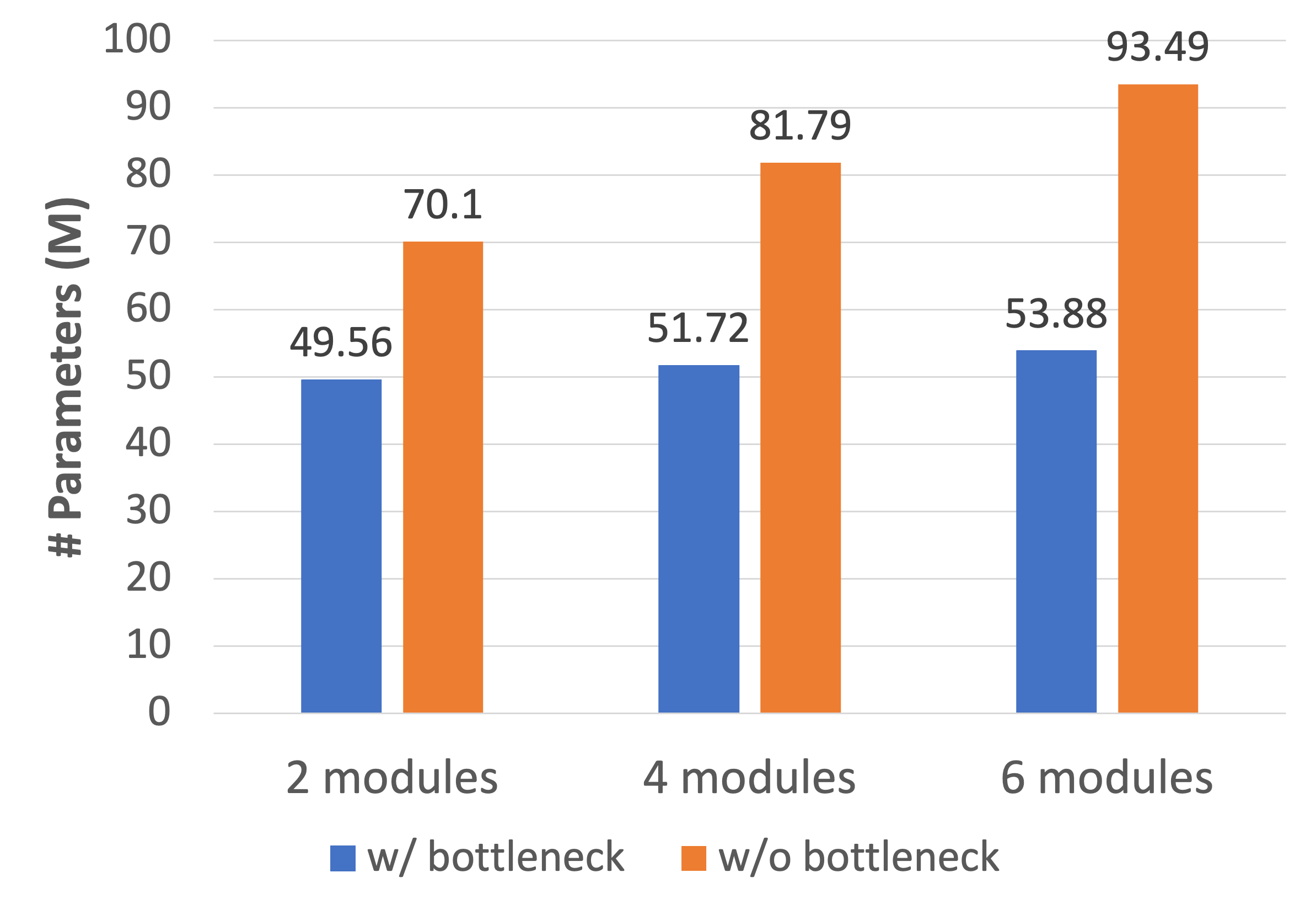}
         % \caption{Effect on the number of parameters}
     \end{subfigure}
     \hfill
     \begin{subfigure}[b]{0.495\textwidth}
         \centering
         \includegraphics[width=\textwidth]{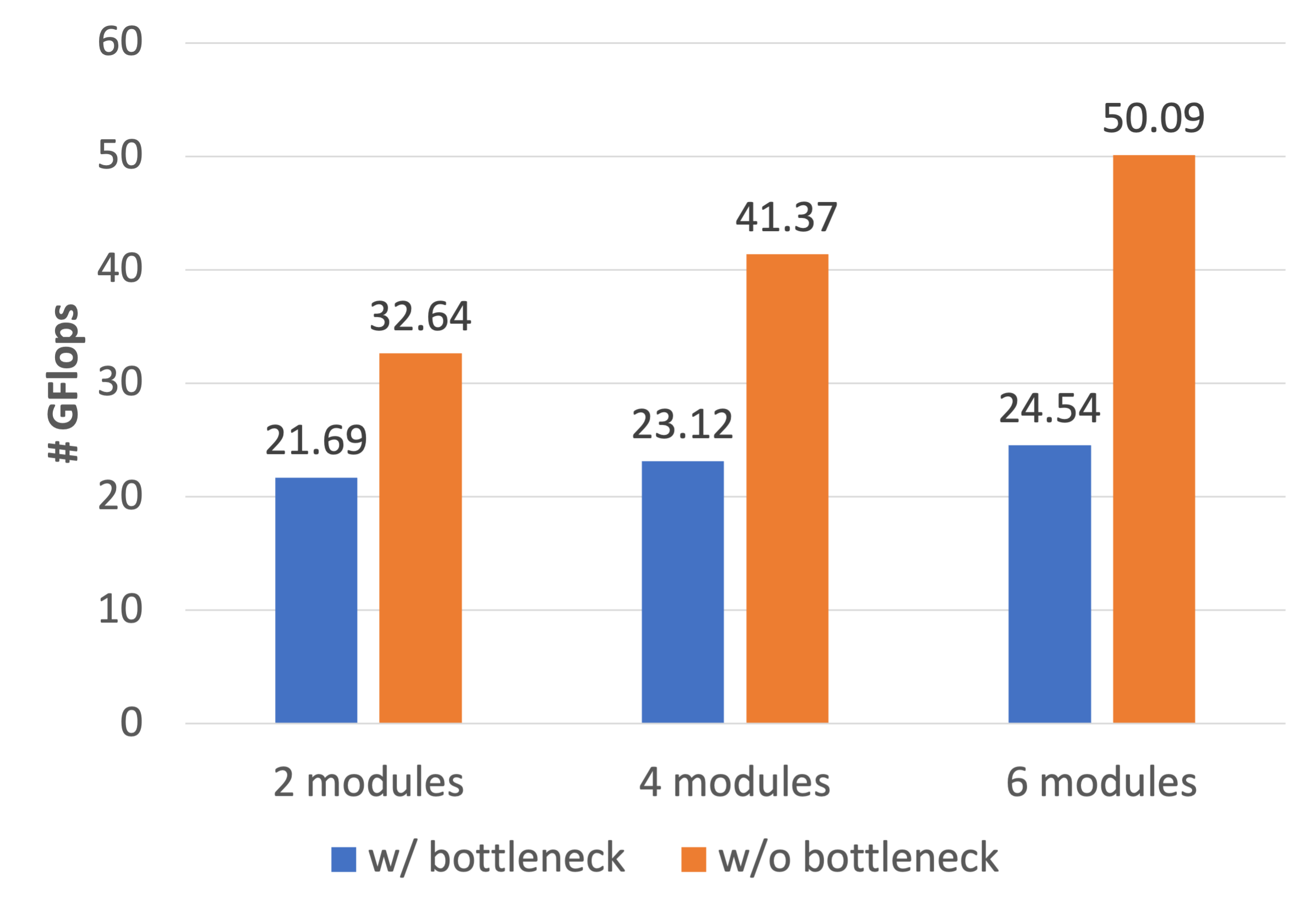}
         % \caption{Effect on the number of GFlops}
     \end{subfigure}
     \caption{Effect of repeating RaBiFPN modules on the number of parameters and GFlops of RaBiT.}
     \label{fig:params_flops}
\end{figure}

Fig.~\ref{fig:params_flops} shows that the bottleneck module reduces the number of parameters and GFlops by about half compared to the same RaBiT network without using it. The bottleneck module also allows the repetition of RaBiFPN blocks without a negligible increase in the number of parameters and GFlops of the network. 

\begin{table*}[!ht]
\caption{Number of parameters and GFLOPs of different methods}
\centering
{\renewcommand{\arraystretch}{1.2}
\begin{tabular}{c|C{2.8cm} C{2cm}}
\hline
\textbf{Method} & \textbf{Parameters (M)} & \textbf{GFLOPs}\\
\hline
\hline
PraNet \cite{pranet}  & 32.55 &   13.11        \\
HarDNet-MSEG \cite{hardnet_mseg} & 33.34 & 11.38         \\
CaraNet \cite{caranet} & 46.64 & 21.69       \\
TransUNet \cite{transunet} & 105.5 & 60.75         \\
%TransFuse-S & - & -         \\
%TransFuse-L & - & -         \\
TransFuse-L* \cite{transfuse} & - & -          \\

SegFormer-B3 \cite{segformer} & 47.22 & 33.68 \\
\hline
%RaBiT 0 block bottleneck & 47.41 & 20.27 \\
%RaBiT 2 block bottleneck & 49.56 & 21.69 \\

RaBiT w/ bottleneck & 51.72 & 23.12 \\
%RaBiT 6 block bottleneck & 53.88 & 24.54 \\
% \hline
%RaBiT 3blocks & 52.59 & 18.02 \\
%RaBiT 4blocks & 53.04 & 23.67 \\
%RaBiT 0 block w/o bottleneck & 58.40 & 23.92         \\

%RaBiT 2 block w/o bottleneck & 70.10 & 32.64        \\

RaBiT w/o bottleneck & 81.79 & 41.37         \\

%RaBiT 6 block w/o bottleneck & 93.49 & 50.09         \\
\hline
\end{tabular}
}
\label{tab:complexity}
\end{table*}

\subsection{Ablation Studies}
\label{sec:ablation_studies}

\textbf{Effectiveness of the RaBiFPN and the bottleneck modules.} We evaluated the effectiveness of these modules through the generalization capability of the models on unseen data. We followed the setup for the first experimental group in Section~\ref{subsec:settings} to compare RaBiT with SegFormer-B3 \cite{segformer}, which employs the original weighted BiFPN module \cite{tan2020efficientdet}. Both models leverage the same MiT-B3 backbone as the encoder. %Notably, the RaBiFPN module in RaBiT can either make use of or exclude the bottleneck module.

\begin{table*}[!ht]
\caption{Ablation study on the effectiveness of the RaBiFPN and bottleneck modules on the unseen data.}
\centering
% \begin{adjustbox}{width=1.1\textwidth}
\def\arraystretch{1.0}
\begin{tabular}{C{1.8cm} C{1.8cm} C{1.8cm}|cc|cc|cc}
\hline
\multirow{2}{*}{wBiFPN} & \multirow{2}{*}{RaBiFPN} & \multirow{2}{*}{Bottleneck}  & \multicolumn{2}{c|}{\textbf{CVC-ColonDB}} & \multicolumn{2}{c|}{\textbf{CVC-T}} & \multicolumn{2}{c}{\textbf{ETIS-Larib}}  \\
\cline{4-9}
                 & & & mDice & mIoU                & mDice & mIoU                   & mDice & mIoU              \\
\hline
\checkmark  & & &  0.812 & 0.733 & 0.900 & 0.829  & 0.784 & 0.708 \\
%SegFormer-Uper & 0.921 & 0.869 & 0.928 & 0.881 & 0.795 & 0.718 & 0.904 & 0.838  & 0.782 & 0.704 \\
& \checkmark  & &   \underline{\textit{0.820}} & \underline{\textit{0.746}} & \textbf{0.905} & \textbf{0.840}  & \textbf{0.828} & \textbf{0.753} \\
% SegFormer-Uper-ARA  & 0.921 & 0.871 & 0.928 & 0.881 & \textbf{0.815} & \textbf{0.737} & \textbf{0.912} & \textbf{0.848}  & 0.787 & 0.708 \\

& \checkmark  & \checkmark   & \textbf{0.824} & \textbf{0.749} & \textbf{0.905} & \underline{\textit{0.839}} & \underline{\textit{0.823}} & \underline{\textit{0.747}} \\[2pt]
\hline
\end{tabular}
% }
% \end{adjustbox}
\label{tab:ablation}
\end{table*}

\begin{table*}[!ht]
\caption{Evaluation metrics for different types of the RA module.}
\centering
% \begin{adjustbox}{width=1.0\textwidth}
\def\arraystretch{1.0}
\begin{tabular}{c|c|C{1.3cm} C{1.3cm} | C{1.3cm} C{1.3cm}  | C{1.3cm} C{1.3cm}}
\hline
Dataset & Activation & Dice$_{seg}$ & IoU$_{seg}$              & Dice$_{non}$ & IoU$_{non}$                & Dice$_{neo}$ & IoU$_{neo}$\\ % \multicolumn{6}{c|}{Neo-Large} & Neo-Small \\ % & \multicolumn{2}{c} {All five datasets}  \\
%\cline{3-9} 
%&type& Dice Seg & IoU Seg              & Dice Non & IoU Non                & Dice Neo & IoU Neo               \\ %& mDice & mIoU            \\
\hline

\multirow{2}{*}{NeoPolyp-Large}  & sigmoid         & 93.66 & 88.08 & 75.18 & 60.23 & 91.34 & 84.07  \\ %& 0.812 &	0.737 \\
 & softmax          & \textbf{94.00} & \textbf{88.59} & \textbf{76.47} & \textbf{61.90} & \textbf{91.65} & \textbf{84.59}   \\ %& 0.835 &	0.763 \\
  %& \textbf{0.839} &	\textbf{0.768}\\

\hline
\multirow{2}{*}{NeoPolyp-Small}  & sigmoid         & 93.40 & 87.61 & 70.22 & 54.10 & 92.39 & 85.85  \\ %& 0.812 &	0.737 \\
 & softmax          & \textbf{94.01} & \textbf{88.69} & \textbf{76.65} & \textbf{62.15} & \textbf{93.80} & \textbf{88.33}   \\ %& 0.835 &	0.763 \\
  %& \textbf{0.839} &	\textbf{0.768}\\

\hline

\end{tabular}
% }
% \end{adjustbox}
\label{tab:typeRA}
\end{table*}

Table~\ref{tab:ablation} shows the superior performance of the RaBiFPN module compared to the original BiFPN module. The mIoU improvement on the CVC-ColonDB and CVC-T datasets, which contain low-resolution images, is above $1\%$. However, the improvement on ETIS-Larib, containing high-resolution images, is up to $4.5\%$ in mIoU. This observation emphasizes the increasing effectiveness of RaBiFPN as image size grows. Although the bottleneck module negligibly impacts performance for large images in ETIS-Larib, it slightly improves the performance on smaller images in CVC-ColonDB. Moreover, the bottleneck module also reduces our network's parameters from 81.79M to 51.72M and decreases computational complexity from 41.37 GFlops to 23.12 GFlops.

%Results are shown in the first two rows of Table~\ref{tab:ablation}. Both network versions show similar metric values across the test datasets, with slight variations of roughly $1\%$. However, one can see from Table~\ref{tab:ablation} that MiTB3 with weighted BiFPN only achieved 0.784 on the ETIS dataset. Meanwhile, the RaBiT with or without bottleneck increases the mDice by roughly 4$\%$. The bottleneck module not only improves results on most datasets but also reduces the parameters from 81.79M down to 51.72M, and reduces the number of flops from 41.37 GFLOPS to 23.12 GFLOPs.

\textbf{Effectiveness of the softmax RA module.} We evaluated the performance of RaBiT using the sigmoid and softmax RA modules on the Neo-Small and Neo-Large datasets. Results in Table~\ref{tab:typeRA} show a consistent improvement when using softmax in the RA module for the multi-class polyp segmentation task. Overall, RaBiT with softmax RA improves results according to all metrics on all datasets. Especially, the Dice$_{non}$ metric improves by 6.43$\%$ on the Neo-Small dataset. This observation suggests the increasing effectiveness of the softmax RA when dealing with small objects such as non-neoplastic polyps.

\subsection{Ablation Study on the Number of RaBiFPN Modules}
We evaluate the performance of RaBiT with 2 - 4 - 6 RaBiFPN modules. Results are shown in Table~\ref{tab:backbone}. Overall, RaBiT with 4 RaBiFPN modules achieved the best result on most datasets except for the ETIS-Larib that contains high-resolution images. We also use RaBiT with 4 RaBiFPN modules in all other experiments in this paper.
\begin{table*}[ht!]
\caption{Evaluation metrics for different numbers of RaBiFPN modules. All results are averaged over five runs.}
\centering
{\renewcommand{\arraystretch}{1.2}
\begin{tabular}{c|cc|cc|cc|cc|cc}
\hline
$\textbf{\#}$ \textbf{RaBiFPN} & \multicolumn{2}{c|}{\textbf{Kvasir}} & \multicolumn{2}{c|}{\textbf{ClinicDB}} & \multicolumn{2}{c|}{\textbf{ColonDB}} & \multicolumn{2}{c|}{\textbf{CVC-T}} & \multicolumn{2}{c} {\textbf{ETIS-Larib}} \\ % & \multicolumn{2}{c} {All five datasets}  \\
\cline{2-11} 
\textbf{modules}& mDice & mIoU               & mDice & mIoU                 & mDice & mIoU                & mDice & mIoU                   & mDice & mIoU  \\ %& mDice & mIoU            \\
\hline
\hline

% 0         & \underline{\textit{0.926}} & \underline{\textit{0.878}} & 0.932 & 0.885 & \underline{\textit{0.823}} & \underline{\textit{0.746}} & \textbf{0.905} &\textbf{ 0.839}  & 0.820 & 0.742 \\

2         & \underline{\textit{0.926}} & \underline{\textit{0.878}} & 0.933 & 0.887 & 0.822 & \textbf{0.749} & \underline{\textit{0.902}} &\underline{\textit{ 0.836}}  & \underline{\textit{0.824}} & 0.746 \\ %& 0.812 &	0.737 \\
4         & \textbf{0.927} & \textbf{0.879} & \textbf{0.936} & \textbf{0.890} & \textbf{0.824} & \textbf{0.749} & \textbf{0.905} & \textbf{0.839}  & 0.823 & \underline{\textit{0.747}} \\ %& 0.835 &	0.763 \\
6          & 0.922 & 0.872 & \underline{\textit{0.935}} & \underline{\textit{0.888}} & 0.822 & 0.745 & \underline{\textit{0.902}} & \underline{\textit{0.836}} & \textbf{0.827} & \textbf{0.750} \\ %& \textbf{0.839} &	\textbf{0.768}\\

\hline
\end{tabular}
}
\label{tab:backbone}
\end{table*}

\section{Conclusion}
\label{sec:conclude}
This paper proposes RaBiT, a new Transformer-based network for automatic and accurate segmentation of colon polyps. RaBiT incorporates a hierarchically structured lightweight Transformer in the encoder and a stack of bidirectional feature pyramid modules in the decoder with repeated lightweight RA modules for iterative polyp boundary refinement. Our method outperforms existing methods across several benchmark datasets for polyp segmentation while maintaining low computational complexity. Furthermore, our method demonstrates high generalization capability in cross-dataset experiments.

In future works, %we will investigate lightweight or sparse self-attention layers to reduce computational complexity. In addition, 
we will exploit alternative feature aggregation mechanisms to improve the representation of high-level semantic features.

%Our model leverages both the advantages of Transformers and CNNs architectures to learn a powerful multi-scale hierarchical feature representation. We also enhance the RA with axial attention by relaxing it with a residual connection. The refinement module allows the network to incrementally correct the polyp boundary from a coarse global map produced by the decoder. Our extensive experiments show that RaBiT significantly outperforms existing state-of-the-art models on popular benchmark datasets.

% \section{Acknowledgments}
% This work was funded by Vingroup Innovation Foundation
% (VINIF) under project code VINIF.2020.DA17.

\bibliographystyle{unsrt}
\bibliography{arxiv}

\section{Appendices}

\subsection{Details of Datasets}
\subsubsection{Binary polyp segmentation.} 
We conducted experiments for binary polyp segmentation on the two following datasets:

\textbf{Kvasir dataset} \cite{kvasir} collected using endoscopic equipment at Vestre Viken Health Trust (VV), Norway. Images were carefully annotated and verified by experienced gastroenterologists from VV and the Cancer Registry of Norway. The dataset consists of 1000 images with different resolutions from $720\times576$ to $1920\times1072$ pixels.

\textbf{CVC-ClinicDB dataset} \cite{cvc_clinic} a database of frames extracted from colonoscopy videos. The dataset consists of 612 images with a resolution of $384\times288$ pixels from 31 colonoscopy sequences. The dataset was used in the training stages of the MICCAI 2015 Sub-Challenge on Automatic Polyp Detection Challenge in Colonoscopy Videos.

\textbf{CVC-ColonDB dataset} \cite{cvc_colon} is provided by the Machine Vision Group (MVG). The dataset consists of 380 images with a resolution of $574\times500$ pixels from 15 short colonoscopy videos.

\textbf{CVC-T dataset} \cite{endo} is the test set of a more extensive dataset called Endoscene. CVC-T consists of 60 images obtained from 44 video sequences acquired from 36 patients.

\textbf{ETIS-Larib dataset} \cite{etis_larib} contains 196 high resolution ($1226\times996$) colonoscopy images.

\subsubsection{Multi-class polyp segmentation.} We conducted experiments for multi-class polyp segmentation on the two following datasets:

\textbf{NeoPolyp-Small} \cite{neopolyp-small} is a public dataset in a Kaggle competition. The dataset contains 1200 images. The training set consists of 1000 images, and the test set consists of 200 images. 

\textbf{NeoPolyp-Large} \cite{ngoc2021neounet} is a bigger version of the NeoPolyp-Small. The training set has 5277 images, and the test set has 1353 images.

\subsection{Qualitative Comparison for Multi-class Polyp Segmentation}
Fig.~\ref{fig:neo} shows more examples of the qualitative comparison between our approach and other state-of-the-art methods on the NeoPolyp-Large dataset. One can see that RaBiT yields more precise segmentation results than other state-of-the-art.
\begin{figure*}[!ht]
\centering
\includegraphics[width=1.0\textwidth]{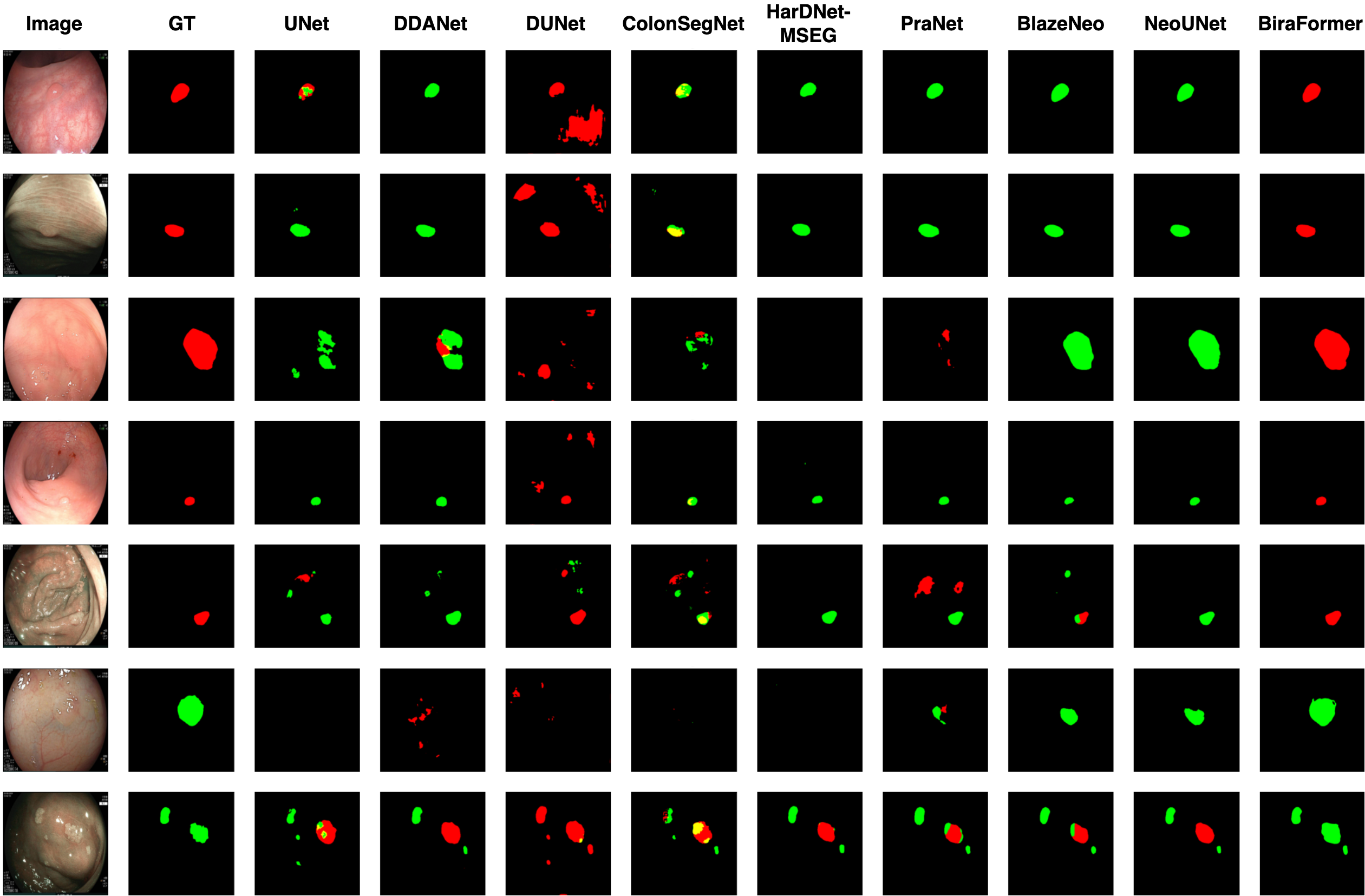}
\caption{Qualitative result comparison in Neo-Large dataset.}
\label{fig:neo}
\end{figure*}

\subsection{Details of Data Augmentation}
\subsubsection{Binary polyp segmentation}
Each image has a probability of 50\% to be augmented by the sequence of the following transforms:
\begin{itemize}  
\item  Randomly rotate the image by 90 degrees;
\item Flip the image either horizontally, vertically, or both horizontally and vertically;
\item  Randomly change the hue, saturation and value of the image;
\item  Randomly change the brightness and contrast of the image;
\item  Blur the image using a Gaussian filter;
\item  Transpose the image by swapping rows and columns;
\item  Random crop or center crop with window size (224,224). 
\end{itemize}

The probability of applying each of these transforms is 50\%, except for the crop transform, which has a probability of 20\%. All augmented images are then resized back to the input size of 384x384 after the augmentation steps. Below are the codes for data augmentation using the \textbf{albumentations} python library:
\begin{lstlisting}
image_transform = Compose([
            rotate.RandomRotate90(),
            transforms.Flip(),
            transforms.HueSaturationValue(),
            transforms.RandomBrightnessContrast(),
            transforms.GaussianBlur(),
            transforms.Transpose(),
            OneOf([
                crop.RandomCrop(224, 224, p=1),
                crop.CenterCrop(224, 224, p=1)
            ], p=0.2),
            resize.Resize(HEIGHT, HEIGHT)
        ], p=0.5)
\end{lstlisting}
\subsubsection{Multi-class polyp segmentation}
Polyps in the multi-class polyp segmentation datasets, especially non-neoplastic ones, seem much smaller, and strong augmentation can negatively affect the results. For that reason, we used lighter augmentation transforms as follows:
\begin{itemize}
\item Blur the image using a Gaussian filter, an average filter, a median filter, or using motion blur;
\item Sharpen the image;
\item Add Gaussian noise to the image;
\item Adjust the values of each color channel using addition or multiplication;
\item Increase or decrease the contrast of the image.
\end{itemize}
All augmented images are then resized back to the input size of 384x384 after the augmentation steps. Below are the codes for data augmentation using the \textbf{imgaug} python library:

\begin{lstlisting}
from imgaug import augmenters as iaa
aug_pipe = iaa.Sequential(
   [    
    iaa.SomeOf((0, 3),
      [
        iaa.OneOf([
           iaa.GaussianBlur((0, 1.5)), 
           iaa.AverageBlur(k=(2,5)), 
           iaa.MedianBlur(k=(3, 5)), 
        ]),
        iaa.Sharpen(alpha=(0, 0.02), lightness=(0.95, 1.05)), 
        imgaug.augmenters.blur.MotionBlur(k=(3, 7), angle=(0, 360)),
        iaa.AdditiveGaussianNoise(loc=0, scale=(0.0, 0.02*255), 
                                  per_channel=0.5), 
         
        iaa.Add((-5, 5), per_channel=0.5), 
        iaa.Multiply((0.95, 1.05), per_channel=0.5), 
        iaa.ContrastNormalization((0.95, 1.05), per_channel=0.5), 
      ],
      random_order=True
    )
  ],
  random_order=True
)

image_datagen_args = {
    'shear_range': 0.1,
    'zoom_range': 0.2,
    'width_shift_range': 0.25,
    'height_shift_range': 0.25,
    'rotation_range': 180,
    'horizontal_flip': True,
    'vertical_flip': True,
    'fill_mode':'constant'
}

image_datagen = ImageDataGenerator(**image_datagen_args)

def augment(image,mask):
  image = image.astype(np.uint8)
  
  if random.random()<0.5:
    seed = random.randint(0,1000000000)
    params = image_datagen.get_random_transform(image.shape,seed = seed)
    image = image_datagen.apply_transform(image, params)
    params = image_datagen.get_random_transform(mask.shape,seed = seed)
    mask = image_datagen.apply_transform(np.expand_dims(mask,-1), 
                                         params)[:,:,0]
                                  
  if random.random()<0.5:
    image = aug_pipe.augment_image(np.array(image).astype(np.uint8)) 
  return image.astype(np.float32),mask
\end{lstlisting}

\end{document}